\pdfoutput=1 
\documentclass[letterpaper, 10 pt, conference]{ieeeconf}  

\IEEEoverridecommandlockouts                              
\usepackage[utf8]{inputenc}
\usepackage[T1]{fontenc}
\usepackage{mathtools, multicol}
\usepackage{amsmath, amsfonts, amssymb, upgreek}
\usepackage{algorithm,algorithmic}
\usepackage{subfig}
\usepackage{hyperref}

\title{\LARGE \bf
Least-Restrictive Multi-Agent Collision Avoidance via Deep Meta Reinforcement Learning and Optimal Control
}

\author{Salar Asayesh$^{1}$, Mo Chen$^{1}$, Mehran Mehrandezh$^{2}$ and Kamal Gupta$^{1}$
\thanks{*This work was not supported by any organization}
\thanks{$^{1}$Faculty of Applied Science,
        Simon Fraser University,Burnaby, BC V5A 1S6, Canada, 
        {\tt\small \{sasayesh, mochen, kamal\} at sfu.ca}}%
\thanks{$^{2}$Faculty of Engineering and Applied Science, University of Regina,
        Regina, SK S4S 0A2, Canada
        {\tt\small mehran.mehrandezh at uregina.ca}}%
}
\renewcommand{\baselinestretch}{.91}

\begin{document}

\maketitle
\thispagestyle{empty}
\pagestyle{empty}

\begin{abstract}

Multi-agent collision-free trajectory planning and control subject to different goal requirements and system dynamics has been extensively studied, and is gaining recent attention in the realm of machine and reinforcement learning. However, in particular when using a large number of agents, constructing a least-restrictive collision avoidance policy is of utmost importance for both classical and learning-based methods. In this paper, we propose a Least-Restrictive Collision Avoidance Module (LR-CAM) that evaluates the safety of multi-agent systems and takes over control only when needed to prevent collisions. The LR-CAM is a single policy that can be wrapped around policies of all agents in a multi-agent system. It allows each agent to pursue any objective as long as it is safe to do so. The benefit of the proposed least-restrictive policy is to only interrupt and overrule the default controller in case of an upcoming inevitable danger. We use a Long Short-Term Memory (LSTM) based Variational Auto-Encoder (VAE) to enable the LR-CAM to account for a varying number of agents in the environment. Moreover, we propose an off-policy meta-reinforcement learning framework with a novel reward function based on a Hamilton-Jacobi value function to train the LR-CAM. The proposed method is fully meta-trained through a ROS based simulation and tested on real multi-agent system. Our results show that LR-CAM outperforms the classical least-restrictive baseline by 30 percent. In addition, we show that even if a subset of agents in a multi-agent system use LR-CAM, the success rate of all agents will increase significantly.
\end{abstract}

\section{INTRODUCTION}
    With increasingly widespread use of autonomous vehicles such as mobile robots and unmanned aerial vehicles (UAVs), control and trajectory planning for multi-agent systems is gaining attention within the scientific community. Especially considering the current pandemic situation, the use of multiple autonomous vehicles in Giga-Factories and hospitals, where physical distancing prevents human workers from performing their regular jobs, is essential. In all these applications, multiple agents need to operate in the same work space carrying out their respective tasks, while not interfering with others. Many of these multi-agent systems are also safety critical. 
    \begin{figure}%
        \centering
        \subfloat[\centering Trajectory visualization on real implementation - colours became lighter as time goes]{{\includegraphics[width=3.5cm]{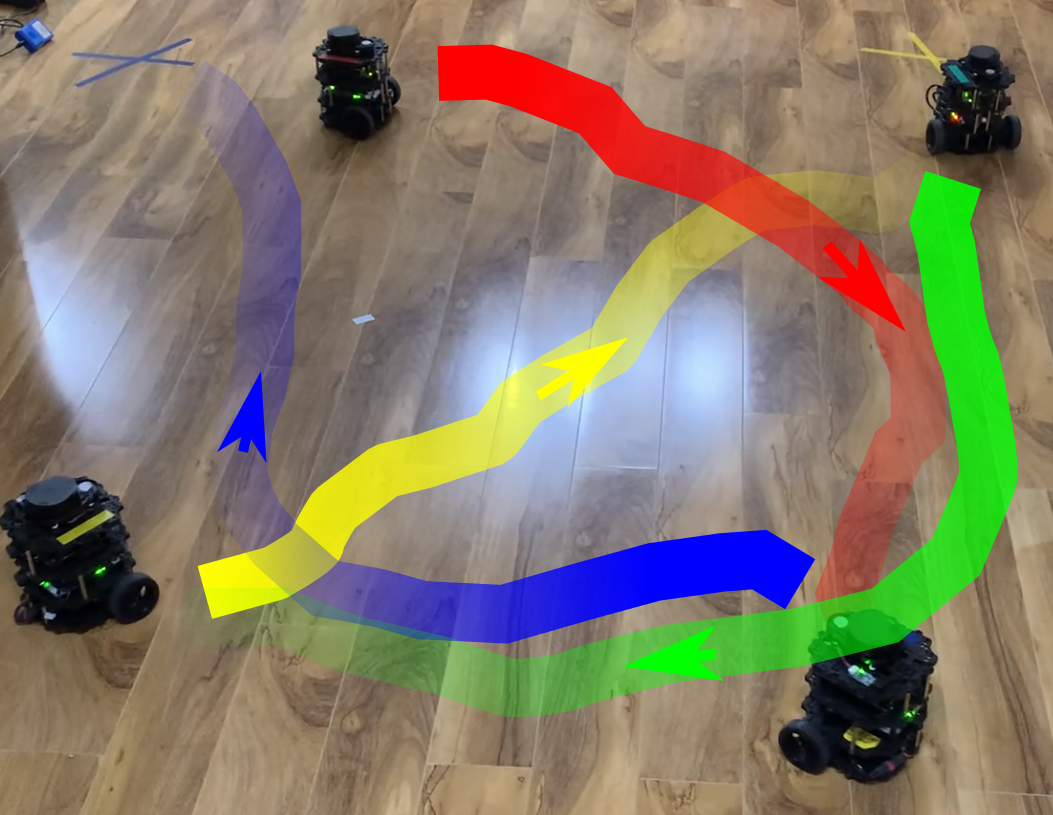} }}%
        \subfloat[\centering Generated trajectory for a 4-agent task through real implementation]{{\includegraphics[width=4.2cm]{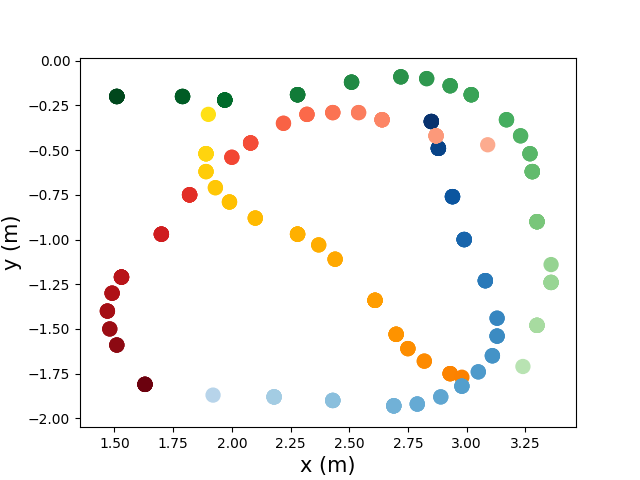} }}%
        \caption{The visualization of generated trajectory in a real-world experiment.}%
        \label{fig:lrcam_overview}%
        \vspace{-7mm}
    \end{figure}

  Recent deep reinforcement learning (RL) methods has produced a quantum leap in solving difficult tasks. Successful applications include video games~\cite{mnih2015human, silver2016mastering}, continuous control~\cite{continuous, trust}, and manipulation tasks from high-dimensional image observations~\cite{levine2018learning, deep}. However, there are still some key challenges in end-to-end training of precise controllers based on sensory observations using deep RL (DRL), which prevents their use in some real-world applications. Moreover, DRL tends to need lots of training episodes to learn a single task, and it is usually not trivial to transfer the learned policy to other similar tasks. Even though recent deep meta-RL methods are capable of adapting to new similar tasks, they usually are on-policy and require massive amount of data from different tasks~\cite{finn2017model,duan2016rl,mishra2017simple}.
    
    Our motivation is to design a Least-Restrictive Collision Avoidance Module (LR-CAM) that can be added on top of any autonomous agent conducting tasks in an environment that consists of multiple other agents. Our approach adds a layer of safety to all agents co-existing in a shared environment, each of whom typically operates based on some default objective/controller. The least-restrictive policy has several benefits compared to other ``do-it-all'' policies that try to both achieve a goal as well as maintain safety. 
    For example, do-it-all policies can require a lot of training data and resources to learn the both objective and safety, and therefore unable to maintain both performance and safety at the same time.
    In addition, if the goal or objective changes, a new policy or complicated meta-trained policy needs to be trained. 
    In contract, the least-restrictive policy enables agents to have any classical or learning based default controller such as goal-reaching controller while providing a higher level of safety for all agents. To achieve this, we propose a distributed control approach using a single deep policy network for all agents in which the LR-CAM of each agent, based on the agent's observations, decides whether it is safe to follow its default controller. If not safe, LR-CAM takes control of the agent and attempts to resolve potential conflicts that may lead to collisions. In our algorithm, first a Long Short Term Memory (LSTM)~\cite{lstm} variational auto-encoder (VAE) maps variable-length observations from a varying number of observed agents into a fixed-length latent space. Then our meta-trained policy supervises the navigation based on the latent space. To remedy the need for a complicated hand-designed reward function, we propose a novel reward function based on Hamilton-Jacobi (HJ) reachability theory \cite{mitchell2005time} to help LR-CAM effectively learn when to take control over each agent.
    
    The key contributions of the proposed work are as follows: (1) We propose LR-CAM, which encompasses any goal-oriented policy of autonomous vehicles to monitor whether collision avoidance maneuvers are needed and if so, takes over control to perform the avoidance maneuvers.
    (2) We propose a model-free off-policy meta-RL algorithm for training the LR-CAM; through the use of an LSTM-VAE, we map observations to a latent space to enable LR-CAM to be used in different environments with a varying number of agents. 
    (3) We propose a novel reward function based on a safety value function computed from HJ reachability. We conduct our training and experiments within a ROS-based controller. We show that the LR-CAM outperforms the baseline classical approach.

\section{Related works}
\label{sec:related_works}
\textbf{Multi-agent collision avoidance}  Research pertinent to the multi-agent collision-free trajectory planning is vast. The authors in \cite{doi:10.1177/027836499801700706}, proposed using dynamic obstacle method in collision avoidance problems; here, the assumed dynamics are used to predict agents' future behaviours. Potential field trajectory planning is another approach~\cite{olfati2002distributed, chuang2007multi} that achieves collision avoidance by slightly changing the trajectory using a repulsive force between vehicles. The authors in \cite{voronoi} proposed the use of buffered Voronoi cells in which each robot computes its Voronoi cells and re-plans its trajectory within a receding horizon to safely reach a specified goal. They assume that the dynamics of the agents are known a priori. In another work, \cite{van2011reciprocal}  proposed an Optimal Reciprocal Collision Avoidance (ORCA) method, which is a well-known classical approach for multi-agent collision avoidance. The ORCA achieves collision avoidance by optimizing a related constrained-cost in a short-time horizon. The problem of trajectory planning and collision avoidance is also studied widely in  safety critical systems under the  differential game framework. In \cite{mitchell2005time, margellos2011hamilton, fisac2015reach}, collision avoidance has been studied using HJ reachability theory in small-scale problems such as those found in a two-agent setup; the authors provide safety and goal reaching guarantees. However, HJ analysis becomes intractable as the number of agents increased. The authors in \cite{chen2016multi} proposed using Mixed Integer Programming (MIP) to establish a higher-level logic around pair-wise collision avoidance using HJ-reachability to alleviate the intractability. 

The problem of multi-agent collision avoidance is also addressed in many previous works within the RL research community \cite{chen2017decentralized,8461113,mitpaper}. The authors in \cite{chen2017decentralized} proposed using RL for a two-agent collision avoidance problem. They extend their previous work in~ \cite{mitpaper} and proposed an RL-based method for single agent motion planning among moving objects, treated as obstacles, without any assumptions on the agents' behavior. They used an actor-critic algorithm to train an agent with an LSTM in the first layer of their policy network. In our work, we meta-trained a separate embedding-VAE to alleviate ``representation learning bottleneck``~\cite{shelhamer2016loss} and varying input sizes. The authors in \cite{8461113} proposed a decentralized policy, trained using an actor-critic RL method which is capable of mapping raw sensory observations to control outputs that results in collision free trajectories. In comparison to the existing RL methods, which propose end-to-end solutions that consider both the problem of goal-reaching and avoidance simultaneously, our proposed LR-CAM takes inspiration from HJ reachability by only interrupting the default controller when needed.

\textbf{Meta Reinforcement Learning} Meta RL has gained attention recently because it enables an agent to easily adapt to new tasks with some shared structures. Meta RL comprises of two main stages. In the first stage, a meta policy is trained with a huge amount of data across different tasks and in second stage, the trained policy is adopted to a specific task with relatively few training iterations. However, most of the proposed meta RL algorithms are on-policy, which means that they often need a huge amount of training data in the meta-training phase. In general, Meta-RL methods can be divided into two main categories~\cite{pmlr-v97-rakelly19a}: (1) context-based, and (2) gradient-based. In the context-based approach, the experiences of tasks are mapped into a latent representation and then a conditional policy is used to adapt to previously seen tasks~\cite{duan2016rl,mishra2017simple, wang2016learning}. The authors in \cite{pmlr-v97-rakelly19a} proposed an off-policy context-based approach by representing the context using probabilistic latent variables. Comparing to our method, it needs an additional latent representation layer to embed the varying dimension of inputs to the latent state, which could make the training procedure inefficient. In gradient-based methods, a policy network~\cite{finn2017model, stadie2018some} or a loss function~\cite{sung2017learning} is meta-trained to capture the task experience in model parameters. Our method can be considered as a gradient-based method built on top of the Model Agnostic Meta Learning (MAML) by~\cite{finn2017model}; however, we meta-train using an off-policy algorithm to improve sample efficiency. 

\section{Preliminaries}

\subsection{Meta Reinforcement Learning}
We define a set of tasks where each task $\mathcal{T}_i$ contains $N=3,\ldots$ or $6$ agents interacting in the same environment and each agent has a different objective with a different default controller. We assume each agent is a Markov Decision Process (MDP) defined by a set of states $\mathcal{X}_i$, a discrete set of actions $\mathcal{U}_i$, initial state distribution $p(x_{i,0})$, transition function $p_i(x_{i,{t+1}}|x_{i,t}, u_{i,t})$, reward function $r_i$ and discount factor $\gamma_i \in (0, 1]$. In addition, we assume each task has a distribution $p(\mathcal{T})$ and is an MDP defined by a tuple $(\mathcal{S}_{\mathcal{T}}, \mathcal{A}_{\mathcal{T}}, P_{\mathcal{T}}, R_{\mathcal{T}}, \Gamma_{\mathcal{T}}, p_{0, {\mathcal{T}}})$ where  $\mathcal{S}_{\mathcal{T}}$ is the task state set, $\mathcal{A}_{\mathcal{T}}$ is the task action set, $P_{\mathcal{T}}(s_{{t+1},\mathcal{T}}|s_{t, \mathcal{T}}, A_{t,\mathcal{T}})$ is the transition function, $ R_{\mathcal{T}}$ is the vector-valued reward function, $\Gamma_{\mathcal{T}} \in \mathbb{R}^N$ is the discount vector and $p_{0, {\mathcal{T}}}$ is the task initial state distribution.
Our goal is to find an optimal policy $\pi(.|s_{t, \mathcal{T}})$ to maximize the sum of expected discounted returns $R_i =  \mathbb{E}[ \sum_{t=t'}^{T}(\gamma_i^{(T-t)}r_{i,t})]$ for all agents across all tasks. We assume reward and transition functions are unknown to the learning agents but can be sampled from the environments.

\subsection{Safety Level Value Function}
In this section we will introduce the notion of safety value function which is used for reward calculation based on HJ theory. We start by assuming each agent operate according to the following approximate ordinary differential equation (ODE) model:
\begin{equation}
    \label{eq:agent_ode}
    \dot{x}_i = f_i(x_i, u_i)
\end{equation}
where $x_i \in \mathcal{X}_i = (p_{x,i}, p_{y,i}, p_{\theta, i})$ and $u_i \in \mathcal{U}_i = (v, \omega)$ are the states and control inputs (linear and angular velocities) of agent $i$ respectively. The dynamic in \eqref{eq:agent_ode} induce the following relative dynamic between each pair of agents:
\begin{equation}
    \label{eq:relative_dynamic}
    \dot{x}_{ij} = g_{ij}(x_{ij}, u_i, u_j), i\neq j
\end{equation}
where $u_i$, $u_j$ are control inputs to agent $i$, $j$ respectively, $x_{ij} = (p_{x, {ij}}, p_{y,{ij}}, p_{\theta, {ij}})$ is the relative states of agent $j$ with respect to agent $i$ and calculated as follows:
\begin{equation}
    \label{relative_coordinate}
        \begin{bmatrix}
        p_{x,ij}\\
        p_{y,ij}\\
        p_{\theta, {ij}}
        \end{bmatrix}
        =\begin{bmatrix}
        \cos{\theta_j} & -\sin{\theta_j} & 0\\
        \sin{\theta_j} & \cos{\theta_j} & 0\\
        0 & 0 & 1
        \end{bmatrix}
        \begin{bmatrix}
        p_{x,i} - p_{x,j} \\
        p_{y,i} - p_{y,j} \\
        p_{\theta,i} - p_{\theta, j}
        \end{bmatrix}\\
\end{equation}

We assume that the functions $g_{ij}$ and $f_i$ are uniformly continuous, bounded and Lipschitz continuous in $x_{ij}$, $x_i$ respectively for fixed $u_i$ and $u_j$. Furthermore, $u_i$ and $u_j$ are chosen from a set of measurable functions \(\mathbb{U}_i\) and \(\mathbb{U}_j\) \cite{chen2016multi}. We define the safety level value function $R_{ij}(t, x_{ij})$:
\begin{equation}
    \label{eq:safety_value_function}
    \mathcal{R}_{ij}(t) = \{x_{ij}:R_{ij}(t, x_{ij})\leq 0\}
\end{equation}

Here, $\mathcal{R}_{ij}$  is the backward reachable set (BRS):
\begin{equation}
\label{eq:brs}
\begin{split}
	   \mathcal{R}_{ij}(t) = \{ x_{ij}:\forall u_i \in \mathbb{U}_i, \exists u_j \in \mathbb{U}_j, x_{ij}(.) \text{  satisfies } \eqref{eq:relative_dynamic}, \\ \exists s\in [0,t], x_{ij} \in \mathcal{D}_{ij} \}
\end{split}
\end{equation}

In \eqref{eq:brs}, $\mathcal{D}_{ij}$ is the danger zone between agents $i$ and $j$:
\begin{equation}
\label{eq:target_set}
\mathcal{D}_{ij} = \{x_i, x_j: (p_{x,i} - p_{x,j})^2 + (p_{y,i} - p_{y,j})^2 \leq d^2\}
\end{equation}
where $d$ is the collision radius. The BRS represents the set of relative states such that agent $i$ will inevitably collide with agent $j$ if agent $j$ applies worst-case control policy to cause a collision. The BRS is calculated by solving the associated HJ partial differential equations by using level set methods \cite{mitchell2005time}. According to \eqref{eq:safety_value_function}, the BRS is the zero sublevel set of the safety value function that encodes the degree of safety based on all relative joint states. Higher is safer, and a negative value means the relative joint states are inside BRS. In this paper we consider the infinite time horizon BRS and safety value function, which can be obtained via taking the limit $t \rightarrow \infty$ as in \cite{chen2016multi}:
\begin{equation}
    \label{eq:time_invariant_value_function}
    R_{ij}(x_{ij}) = \lim_{t\to\infty} R_{ij}(t, x_{ij}).
\end{equation}

\section{Approach}
We are interested in learning the LR-CAM policy that allows all agents to follow their default controllers as long as no intervention is needed for preventing collision. The default controller could be used for goal reaching or simply be given by a human controller. However, when all, or a subset of the agents are in potential conflict with each other, the LR-CAM will intervene and take over control to attempt to allow vehicles to revert back to safety. 
As a result, our policy not only decides whether each agent can follow its default controller or avoid danger, but also detects possible dangers from a latent-space observation.
\subsection{Observation, Action, and Latent Spaces}
We use the relative coordinate system defined in \eqref{relative_coordinate}. Using the relative coordinate not only reduces the dimension of observation space, but also can be extracted using on-board sensors of agents. However, as we will later explain in an ablation study, using information from only the current time step as observation is not enough to learn to predict upcoming dangers efficiently.  As a result, we define the observation based on a history of states and action as follows:
\begin{equation}
\begin{split}
    \label{observation_space_new}
    O_{i, t, \mathcal{T}} = (o_{i, t, \mathcal{T}}, o_{i, t-1, \mathcal{T}}, o_{i, t-2, \mathcal{T}}, o_{i, t-3, \mathcal{T}},\\
    a_{i, t-1}, a_{i, t-2}, a_{i, t-3})
\end{split}
\end{equation}
where $a_{i, t}$ is the action of agent $i$ at time t and $o_{i, t, \mathcal{T}}$ is defined by

\begin{equation}
    \label{observation_in_time_step}
    o_{i, t, \mathcal{T}} = (x_{i1}, x_{i2},\ldots, x_{i,i-1}, x_{i,i+1},\ldots, x_{iN})
\end{equation}
 
The action space of each task MDP with $N$ agents are defines as follows:
\begin{equation}
    \label{task_action_space}
    A_{t, \mathcal{T}} = (a_{t, 0},\;\ldots,\;a_{t, N})
\end{equation}
where the action of each individual agent is defines as
\begin{equation}
\begin{split}
    \label{action_space}
    a_{t,i} \in \mathcal{U}_i = \{0\text{ (default-controller)}, 1\text{ (turning-right)}, \\ 2\text{  (turning-left)}\}.
\end{split}
\end{equation}

In \eqref{action_space}, ``default-controller'' means agent is allowed to execute any default controller such as a goal-reaching controller, and this controller can be different for each agent inside a task. The other two classes of actions refers to avoidance actions that agent should execute to prevent safety violation with the other agents. Taking inspiration from reachability theory \cite{mitchell2005time}, the least restrictive controller lets agents execute their own default controllers unless some upcoming danger is inevitable, in which case agents should avoid danger with their maximum actuation capacity. As a result the avoidance actions are translated to
\begin{equation}
    \begin{split}
    \label{avoidance_actions}
        \text{turning-right} := [v_{max},\;\omega_{max}]\\
        \text{turning-left} := [v_{max},\;-\omega_{max}].
    \end{split}
\end{equation}
where $v$ and $\omega$ are the linear and angular velocities respectively.\\

\begin{figure}
    \centering
    \includegraphics[scale=0.45]{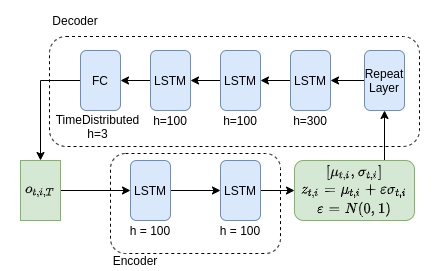}
    \caption{Architecture of LSTM-VAE. An input observation $o_{i, t, \mathcal{T}}$ is converted to the latent space using the encoder network and is sampled via reparametrization trick. A decoder network converts the sampled latent space to the observation an both networks are meta-trained by maximizing the ELBO (\ref{elbo})}
    \label{fig:vae}
    \vspace{-6mm}
\end{figure}

To handle the varying size of observation space in \eqref{observation_space_new} across different tasks, we use an encoder part of a LSTM-based VAE to encode the observation to a fixed-size latent space as is illustrated in Fig. \ref{fig:vae}. The VAE consist of two parts: (1) an LSTM-based encoder network $q_\mathcal{\phi}(z_{i,t}|o_{i, t, \mathcal{T}})$, parametrized by $\mathcal{\phi}$, which encodes the observation $o$ (\ref{observation_in_time_step}) to a fixed size latent space, and (2) an LSTM-based decoder network  $q_\mathcal{\phi}(o_{i, t, \mathcal{T}|z_{i,t}})$ parametrized by $\mathcal{\phi}$. Given latent variables $z_{i,t}$, we define the fixed-size augmented latent state $Z_{i,t}$ as follows:
\begin{equation}
    \label{eq:augmented_latent_space}
    Z_{i,t} = (z_{i, t}, z_{i, t-1}, z_{i, t-2}, z_{i, t-3}, a_{i, t-1}, a_{i, t-2}, a_{i, t-3})
\end{equation}
\subsection{Reward Engineering}
In a multi-agent system with distributed control, where each agent acts independently, each agent needs a notion of safety margin with respect to other agents. While a naive distance function is widely selected as the base metric for setting up a reward function, we instead propose on using a value function derived from the HJ reachability. It encodes the degree of safety as a function of all relative state variables rather than just relative position, as defined in \eqref{relative_coordinate}. 
HJ reachability also accounts for the system's dynamics. 
We calculate the time-invariant safety value function $R_{ij}(x_{ij})$ as described in \eqref{eq:relative_dynamic} to \eqref{eq:time_invariant_value_function} and we define the reward function of agent $i$ as follows:
\begin{equation}
    \label{reward_function}
    r_{i,t,\mathcal{T}}(o_{i,t,\mathcal{T}}, a_{t,i}) = 
    \begin{cases}
    -5  \;\;\;\;\;\text{if } \forall j, R_{ij} \ge 1 \text{ AND } a_{t,i, \mathcal{T}}\neq 0\\
    k\times\min_j(R_{ij})  \;\;\;\;\;\;\;\;\;\text{ if }\forall j, \exists R_{ij} \leq 0\\
    -300 \;\;\;\text{ if collision with any agent } j \\
    300 \;\;\text{finish the task without collision}
    \end{cases}
\end{equation}
where $j=1,...,N, i\neq j$ and $k$ is constant factor. This reward function is derived from the following logic. The first row implies that when there is not any possible danger (all safety values are greater than 1), the agent will receive a big punishment if it does not follow its default controller. The second row implies that if the pairwise safety between the agent and other agents is less than zero, then the reward will be the most negative safety value across all agents $j$ with a scaling factor $k$. The other parts of (\ref{reward_function}) are the sparse rewards that the agent will receive at the end of episode depending on whether it finishes its own task or collides with another agent. We are not considering any continuous reward if all safety values are in the range $(0, 1]$. However, since we are defining discount factor $\gamma$ in our MDP definition, decreasing the discount factor motivate agents to finish their tasks sooner.

\subsection{Off-Policy Meta-Reinforcement Learning}
In this section, we explain how we combine our proposed LSTM-VAE with an off-policy RL algorithm, and how we train them all in a meta-training loop. Our off-policy meta RL makes the learning procedure sample efficient in both training and adaptation loops. However, efficient off-policy meta-RL has two main challenges  \cite{pmlr-v97-rakelly19a}. (1) the same distribution of data should be used for both meta-training and meta-testing. This indicates that since the meta-testing phase is usually done on on-policy data, the meta-training is also should be done on on-policy data as well. (2) The meta policy should be able to reason about the distribution of experience by optimizing the distribution of visited states. This means the policy gradient based methods are more appealing for meta reinforcement learning. To remedy these challenges, we propose using Proximal Policy Optimization (PPO) as the off-policy actor-critic method \cite{ppo} and we meta-train our algorithm using MAML \cite{finn2017model}. The PPO algorithm is not truly off-policy; however, it enables multiple epochs of update from replay buffer and is more sample efficient compared to other on-policy methods. Moreover, directly learning a policy can be more effective on stochastic exploration rather than deriving the policy from a value network, as is done value-based methods.  
The LSTM VAE is meta-trained via MAML \cite{finn2017model} across different tasks using reparameterization trick by maximizing the resulting Evidence Lower Bound (ELBO) \cite{elbo}:
\begin{equation}
\label{elbo}
    \mathbb{E}_{O \sim \mathcal{T}}[\mathbb{E}_{z \sim q}[\log p_\mathcal{\phi}(O|z)] - D_{KL}(q_\mathcal{\phi}(z|O)\;||\;p(z))
\end{equation}
where $p(z)$ is prior distribution over $z$. To prevent LSTM blocks from forgetting critical data, for the observation of each agent $i$ in \eqref{observation_in_time_step}, we sort the $(x_{ij})$ in ascending order of HJ safety value function $R_{ij}$, and feed them to LSTM block such that the relative state of most critical agent -- the agent $j$ with minimum $R_{ij}$ -- is fed last.\\
The critic is meta-trained via the following bootstrap update rule:
\begin{equation}
\label{eq:loss_critic}
\begin{split}
    L_{critic} = \mathbb{E}_{s,r,s' \sim \mathcal{T}, Z \sim q_\phi(z|s) , Z'\sim q_\phi(Z'|s)}[Q_\psi(Z)\\ - (r + \gamma \bar{Q}_\psi(Z'))]^2 
\end{split}
\end{equation}
where $\bar{Q}$ is the target network. The policy which predict the categorical distribution over actions defined in \eqref{action_space} is meta-trained using the following clipped surrogate loss with KL penalty term:
\begin{equation}
    \label{eq:policy_loss}
    \begin{split}
    L_{policy} = \mathbb{E}_{s,r\sim \mathcal{T}, Z\sim q_\phi(Z|s)}[\min(r_t(\theta)\bar{A}_t, \text{clip}(r_t(\theta),1-\epsilon,\\1+\epsilon)\bar{A}_t - \beta D_{KL}(\pi_{\theta_{old}}(.|Z)||\pi_\theta(.|Z))]
    \end{split}
\end{equation}
where the probability ratio \(r_t(\theta) = \frac{\pi_\theta(a_t|z_t)}{\pi_{\theta_{old}}(a_t|z_t)}\), \(\epsilon\) is the clipping hyper parameter value, \(\beta\) is a constant coefficient, \(D_{KL}\) is shorthand of KL-divergence between new and old policy and \(\bar{A}_t\) is the Generalized Advantage Estimation (GAE) which is computed as in~\cite{gae}:
\begin{equation}
    \label{eq:advantage}
    \begin{aligned}
      \bar{A}_t = \delta_t + (\gamma)\delta_{t+1}+...+(\gamma)^{T-t+1}\delta_{T-1}\\
      \text{where} \; \delta_t = r_t(s, a) + \gamma Q_\psi(Z_{t+1}) - Q_\psi(Z_t)\\
    \end{aligned}
\end{equation}
The Alg. \ref{alg:alg1} illustrates the proposed meta-training loop.
\begin{algorithm}
 \caption{proposed meta-training loop}
 \label{alg:alg1}
 \begin{algorithmic}[1]
 \REQUIRE 
  initialize parameters \(\phi, \theta, \psi\)\\
  initialize reply buffers \(\mathcal{B}^i\) for each training task
  \WHILE{True}
  \FOR{k=1, ...,$K$} 
  \FOR{each task \(\{\mathcal{T}_i\}\)}
  \FOR{each agent in task \(\{\mathcal{T}_i\}\)}
  \STATE {$\text{sample} \;z_t \sim q_\phi(\mathbf{z}|s)$}
  \STATE {Calculate $Z$ using (\ref{eq:augmented_latent_space})}
  \STATE {$\text{Roll out policy \(\pi_\theta(a_t|Z_t)\) to}$}
  \text{collect data \{\(o_t, a_t, r_t, o'_t\)\} and}
  \text{add to \(\mathcal{B}^i\)}
  \ENDFOR
  \ENDFOR
  \ENDFOR
  \FOR{each task \(\{\mathcal{T}_i\}\)}
  \STATE {$\text{pull data from \(\mathcal{B}^i\) and calculate GAE}$}
  \text{and update buffer with}
  \text{\{\(o_t, a_t, r_t, o'_t, \bar{A}_t\)\}}
  \ENDFOR
  \FOR{step in PPO off-policy training steps}
  \FOR{\(\mathcal{T}_i\)}
  \STATE{$\text{sample mini batch } b^i \sim \mathcal{B}^i$}
  \STATE{$\text{sample}\; \mathbf{z} \sim q_\phi(z|b^i)$}
  \STATE{Calculate $Z$}
  \IF{step < heat-up-training-steps}
  \STATE{$\phi \leftarrow \phi - lr_1 \nabla_\phi L_{ELBO}(b^i, \mathbf{Z})$}
  \ELSE
  \STATE{$\phi \leftarrow \phi - lr_1 \nabla_\phi L_{ELBO}(b^i, \mathbf{Z})$}
  \STATE{$\psi \leftarrow \psi - lr_2 \nabla_\psi L_{critic}(b^i, \mathbf{Z})$}
  \STATE{$\theta \leftarrow \theta - lr_3 \nabla_\theta L_{actor}(b^i, \mathbf{Z})$}
  \ENDIF
  \ENDFOR
  \ENDFOR
  \ENDWHILE
 \end{algorithmic}
 \end{algorithm}

\section{Simulated and Real-World Experiments}
In this section, we will explain about the implementation, training and performance of our proposed method in both simulation and experimentation. We compare our method with a classical sub-optimal approach and evaluate the specific choice of our design through ablation study. It should be noted that since the LR-CAM objective is to be least restrictive by only interrupting the default controller to avoid danger, it cannot be compared with previous collision avoidance algorithms such as ORCA \cite{van2008reciprocal} that solve the goal reaching and avoidance simultaneously. As a result, we chose the method presented in \cite{chen2016multi} as the classical baseline because similar to LR-CAM it tries to maintain safety by interrupting the default controller. 

\subsection{Implementation}
Since there are no standard benchmark tests set for multi-agent RL, we created our own simulation environments using the Gazebo and ROS open source software. We created four different tasks, with three to six agents working simultaneously.  We considered a default controller for each agents, in all the aforementioned four tasks, with the objective of reaching a specific location within the environment or simply to follow a path. To show the performance of our algorithm in presence of static obstacles in the environment, we make the agents stationary under two scenarios: (1) we randomly freeze some of agents, or (2) we let agents stop-and-stay in the environment when they finish their default tasks. 

In simulation We localize agents based on their odometry data and, in real experiment we localize agents using their on-board sensors. We make the localization information available to all other agents. To make the simulation/training runs a better reflection of the reality, we also add white noises to all states.
The LR-CAM module is implemented in TensorFlow 2.0. 
For real world experiments and simulations, we used the TurtleBot3 Burger robots which are equipped with an onboard 360-degree 2D LiDAR and Gyro (Fig. \ref{fig:lrcam_overview}-a). We also upgrade their raspberry pi development board to Nvidia Jetson Nano to increase their computational capacity. 

To calculate the reward function, we approximate the Turtlebot dynamic with simple Dubins car, where the pairwise relative dynamic of \eqref{eq:relative_dynamic} is written as the following planar kinematic model:
\begin{equation}
\label{eq:relative_dynamic_dubin}
\begin{aligned}
    \dot{p}_{x,ij} &= -v + v\cos(\theta_{ij}) + \omega_i p_{y,ij}\\
    \dot{p}_{y,ij} &= v\sin(\theta_{ij}) - \omega_i p_{x,ij}\\
    \dot{p}_{\theta, ij} &= \omega_j - \omega_i,  |\omega_i|,|\omega_j| \leq \Bar{\omega}
\end{aligned}
\end{equation}

\begin{figure}
    \centering
    \includegraphics[scale=0.25]{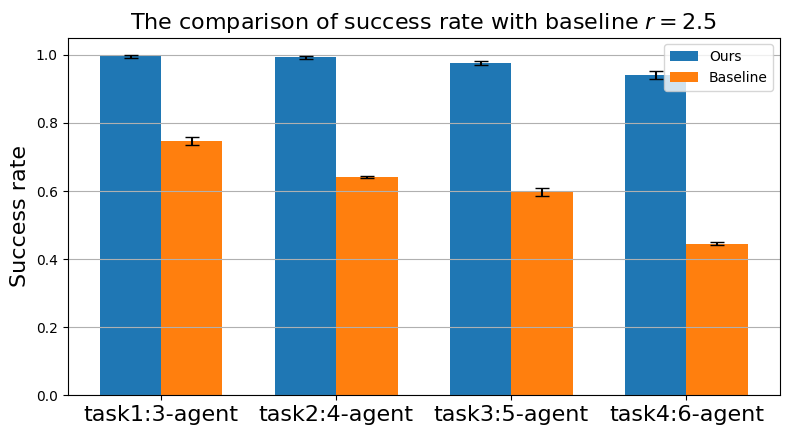}
    \caption{Comparison with Baseline}
    \label{fig:compare_baseline}
    \vspace{-4mm}
\end{figure}
For training and testing, we consider moderate and difficult scenarios. In the former, initial and target locations for each agent are randomly selected between two concentric circles with a radii $r_1$ and $r_2$ ($r_1\leq r_2$), respectively. Then, we add random perturbation to the agents' initial states. Under this setting, one agent usually comes to conflict with maximum two other agents. In the latter, however, agents' locations are randomly initialized around a circle with radius $r$ and their objectives are chosen such that agents would pass through the center of this circle to reach their goal locations. Under this scenario, each agent can be in conflict with two or more agents. To evaluate the performance of our algorithm, we calculate the success rate to be the fraction of agents that succeeded in reaching a goal location without colliding with other agents over 100 trials for each scenario.
\subsection{Results}
\begin{figure}
    \centering
    \includegraphics[scale=0.3]{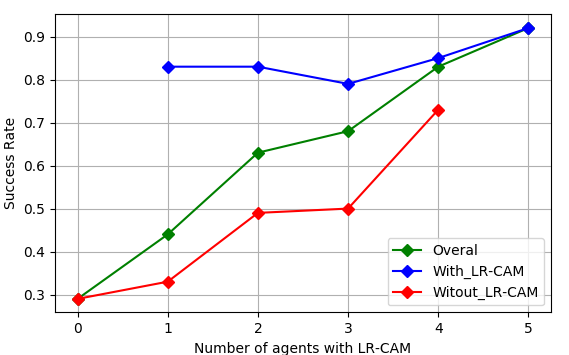}
    \caption{comparison of success rate when a subset of agents in a difficult-task scenario(with $r=1.7$) uses LR-CAM.}
    \label{fig:cpartial_lrcam}
    \vspace{-4mm}
\end{figure}

Fig. \ref{fig:compare_baseline} compares the success rate of our proposed approach with baseline across different tasks in a difficult test case scenario. To calculate the variance of success rate we repeat each success rate calculation 5-times. As the number of agents in a task increased, the success rate is decreased for both algorithms; however, the variation is negligible for our method (more than 90 percent success rate for all cases). In addition, our proposed method outperforms the baseline approach at least by 30 percent in all cases.

we also evaluate the performance of our proposed method in the case in which only some of the agents uses LR-CAM. To this end, we designed the following experiment. First, in a 5-agent task we start with no agent and increase it all the way to all agents using the LR-CAM. Results for each cases in visualized in Fig. \ref{fig:cpartial_lrcam}. The results indicate that even only if some of agents uses LR-CAM, the success rate for agents who use LR-CAM and also for the rest of agents increased significantly. 

We also tested our approach in an unseen environment involving $N=8$ agents. The first row of Table \ref{table:interapolation} shows the success rate in a test case where $N=8$ in \eqref{observation_in_time_step}. In the second row we only feed each agent the six lowest safety values while ignoring the remaining 2 safety values.

\begin{figure}
    \centering
    \includegraphics[scale=0.4]{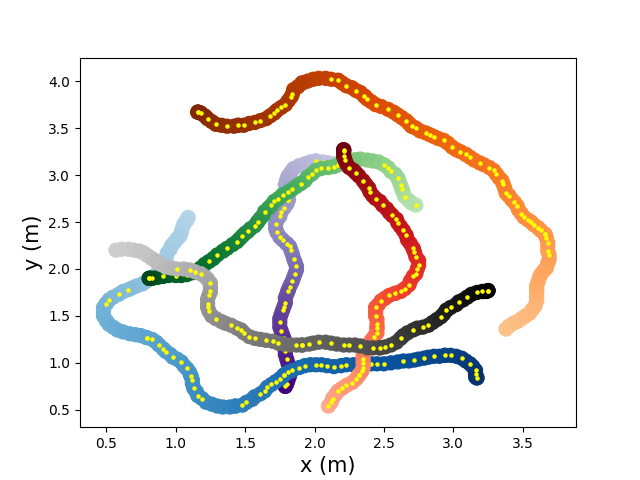}
    \caption{Trajectory visualization for  6-agent task in a crowded initial condition ($r=1.5$). Agents are color coded and the level of opacity in the figures is proportional to the elapsed time. Yellow dots inside tubes indicated and all the time-steps that the LR-CAM of agent is interrupting the default controller.}
    \label{fig:6-agent-trajectory}
    \vspace{-5mm}
\end{figure}

\begin{figure}
    \centering
    \includegraphics[scale=0.4]{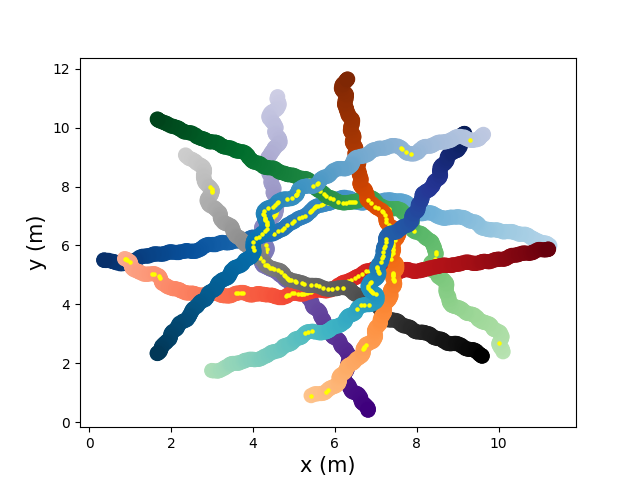}
    \caption{Trajectory visualization for  an unseen 8-agent task in a long range initial condition ($r=5.5$)}
    \label{fig:8-agent-trajectory}
    \vspace{-6mm}
\end{figure}

Fig. \ref{fig:6-agent-trajectory} and Fig. \ref{fig:8-agent-trajectory} show trajectory visualization for a 6-agent task and an 8-agent task respectively. In Fig. \ref{fig:6-agent-trajectory}, we chose a crowded initial condition with $r=1.5$ and in Fig. \ref{fig:8-agent-trajectory}, agents are initialized a bit farther from each other. In both figures, the agents' trajectories are color coded with the opacity proportional to the elapsed time. 

To visualize how exactly LR-CAM intervenes the default controller we overlay yellow dots inside the colored tubes for all time steps in which LR-CAM interrupts the default controller to perform avoidance actions. Based on these trajectories, when the agents are in dangerous configurations, LR-CAM takes over control and ,when the safety value is relatively high, it will let the agent execute its default controller.
Fig. \ref{fig:lrcam_overview}-b shows the trajectories of four TurtleBots in a real-world 4-agent task. 

Additional experiments involving two to four agents can be found at \url{https://bit.ly/34K8YKB} .
\begin{table}[h]
\caption{Performance of LR-CAM in an unseen task with 8 agents}
\label{table:interapolation}
\begin{center}
\begin{tabular}{|c||c|}
\hline
Test Case & Success Rate \\ 
\hline
Observation Space with all agents & 0.85\\ 
\hline
Observation space with first six critical agents & 0.91\\ 
\hline
\end{tabular}
\end{center}
\vspace{-5mm}
\end{table}

\subsection{Ablation Studies}
\textbf{Reward Design.} To demonstrate the benefits of our reward function, we meta-trained a separate policy using a naive distance-function based reward. In this setting we modify \eqref{reward_function} using the Euclidean distance ($L2_{ij}$) instead of safety value function ($R_{ij}$) as follows:
\begin{equation}
    \label{ablation-reward_function}
    \begin{split}
    r_{i,t,\mathcal{T}}(o_{i,t,\mathcal{T}}, a_{t,i}) = \qquad	\qquad	\qquad	\qquad	\qquad	\qquad	\qquad	\qquad	\\
    \begin{cases}
    -5  \;\;\;\;\qquad	\qquad	\qquad	\text{if } \forall i, L2_{ij} \ge 1 \text{ AND } a_{t,i, \mathcal{T}}\neq 0\\
    k\times\min_j((L2_{ij} - d))  \qquad	\qquad	\;\;\;\;\text{ if } \forall j,\; \exists L2_{ij} \leq d\\
    -300 \;\;\;\;\qquad \qquad \qquad \text{ if collision with any agent } j\\
    300 \;\;\;\qquad\qquad\qquad\text{finish the task without collision}
    \end{cases}
    \end{split}
\end{equation}
where $d$ is the collision radius defined in \eqref{eq:target_set}. Table \ref{table:ablation-success-rate} shows the comparison of success rate for using the HJ-based reward (first row) and the distance-based reward (second row) for difficult test scenarios with $r=1.7$. We observed that not only did the success rate improve by approximately 10\%, but also the training time reduced by 25\% when using our proposed HJ-based reward. 
\begin{table}[h]
\caption{Ablation study for reward design - success rate comparison}
\label{table:ablation-success-rate}
\begin{center}
\begin{tabular}{|c||c|c|c|}
\hline
Algorithm & 4-agent & 5-agent & 6-agent \\ 
\hline
Ours with HJ reward & 0.99 & 0.93 & 0.91\\ 
\hline
Ours with naive distance reward & 0.89 & 0.86 & 0.75\\ 
\hline
\end{tabular}
\end{center}
\vspace{-5mm}
\end{table}
To compare how LR-CAM evaluates the safety and upcoming dangers, we define ``restrictiveness factor'' to be the fraction of interrupting actions over all actions in a total of 100 trials for each scenario/task. As it can be seen from Table \ref{table:ablation-restrictiveness}, using our proposed HJ-based reward improved the restrictiveness factor by 10\%, which means that our proposed reward is 10\% less restrictive and allows the default controller to be used more often.
\begin{table}[h]
\caption{Ablation study for reward design - restrictiveness comparison}
\label{table:ablation-restrictiveness}
\begin{center}
\begin{tabular}{|c||c|c|c|}
\hline
Algorithm & 4-agent & 5-agent & 6-agent \\ 
\hline
Ours with HJ reward & 0.46 & 0.45 & 0.48\\ 
\hline
Ours with naive distance reward & 0.54 & 0.54 & 0.48\\ 
\hline
\end{tabular}
\end{center}
\vspace{-5mm}
\end{table}

\textbf{Observation Space Design.} To demonstrate the effectiveness of our observation space design, we trained a policy using only information from a single time step as observation. According to Table \ref{table:ablation-observation}, using a history of observations/actions improved the success rate significantly. 
\begin{table}[h]
\caption{Ablation study for observation space design - success rate comparison}
\label{table:ablation-observation}
\begin{center}
\begin{tabular}{|c||c|c|c|}
\hline
Algorithm & 4-agent & 5-agent & 6-agent \\ 
\hline
Ours with history of observations & 0.99 & 0.93 & 0.91\\ 
\hline
Ours with single step observation & 0.82 & 0.58 & 0.56\\ 
\hline
\end{tabular}
\end{center}
\vspace{-5mm}
\end{table}

\section{CONCLUSIONS}

This paper presented the least-restrictive collision avoidance module (LR-CAM) that can be added on top of autonomous agents to intervene and avoid collisions based on the joint configuration of multiple agents. We also proposed a novel off-policy meta reinforcement learning framework to train the LR-CAM. Future work includes augmenting sensory observations as the main source of input.






\clearpage

\bibliographystyle{IEEEtran}
\bibliography{ref}
\end{document}